\documentclass[10pt,twocolumn,letterpaper]{IEEEtran} 
\usepackage[utf8]{inputenc}
\usepackage{graphicx}
\usepackage{amsfonts}
\usepackage{enumitem}
\usepackage{braket}
\usepackage{amssymb}
\usepackage{xcolor}
\usepackage{braket}
\usepackage{caption}
\usepackage{soul}
\usepackage{subfig}
\usepackage{hyperref}
\usepackage{dblfloatfix}
\usepackage{cite}
\usepackage{graphicx}
\usepackage{textcomp}
\usepackage{physics}
\usepackage{bbold}

\usepackage{algorithm}
\usepackage{algorithmic}

\DeclareCaptionLabelFormat{andtable}{#1~#2  \&  \tablename~\thetable}

\def\ddefloop#1{\ifx\ddefloop#1\else\ddef{#1}\expandafter\ddefloop\fi}
\def\ddef#1{\expandafter\def\csname bb#1\endcsname{\ensuremath{\mathbb{#1}}}}
\ddefloop ABCDEFGHIJKLMNOPQRSTUVWXYZ\ddefloop
\def\ddef#1{\expandafter\def\csname c#1\endcsname{\ensuremath{\mathcal{#1}}}}
\ddefloop ABCDEFGHIJKLMNOPQRSTUVWXYZ\ddefloop
\def\ddef#1{\expandafter\def\csname v#1\endcsname{\ensuremath{\boldsymbol{#1}}}}
\ddefloop ABCDEFGHIJKLMNOPQRSTUVWXYZabcdefghijklmnopqrstuvwxyz\ddefloop
\def\ddef#1{\expandafter\def\csname v#1\endcsname{\ensuremath{\boldsymbol{\csname #1\endcsname}}}}
\ddefloop {alpha}{beta}{gamma}{delta}{epsilon}{varepsilon}{zeta}{eta}{theta}{vartheta}{iota}{kappa}{lambda}{mu}{nu}{xi}{pi}{varpi}{rho}{varrho}{sigma}{varsigma}{tau}{upsilon}{phi}{varphi}{chi}{psi}{omega}{Gamma}{Delta}{Theta}{Lambda}{Xi}{Pi}{Sigma}{varSigma}{Upsilon}{Phi}{Psi}{Omega}{ell}\ddefloop

\def\BibTeX{{\rm B\kern-.05em{\sc i\kern-.025em b}\kern-.08em
    T\kern-.1667em\lower.7ex\hbox{E}\kern-.125emX}}

\begin{document}

\title{Energy-based Generative Models for Target-specific Drug Discovery
\vspace{5mm}
} 

\author{\large Junde Li, Collin Beaudoin, and Swaroop Ghosh \\
 Pennsylvania State University, University Park, PA USA\\
 \{jul1512, cpb5867, szg212\}@psu.edu
 \vspace{5mm}}


\maketitle
\thispagestyle{empty}\pagestyle{empty}

\begin{abstract}
   Drug targets are the main focus of drug discovery due to their key role in disease pathogenesis. Computational approaches are widely applied to drug development because of the increasing availability of biological molecular datasets. Popular generative approaches can create new drug molecules by learning the given molecule distributions. However, these approaches are mostly not for target-specific drug discovery. We developed an energy-based probabilistic model for computational target-specific drug discovery. Results show that our proposed TagMol can generate molecules with similar binding affinity scores as \emph{real} molecules. GAT-based models showed faster and better learning relative to GCN baseline models.
\end{abstract}


\section{\bf Introduction}
\label{sec:intro}

\begin{figure}
\centering
\includegraphics[width=5.5cm]{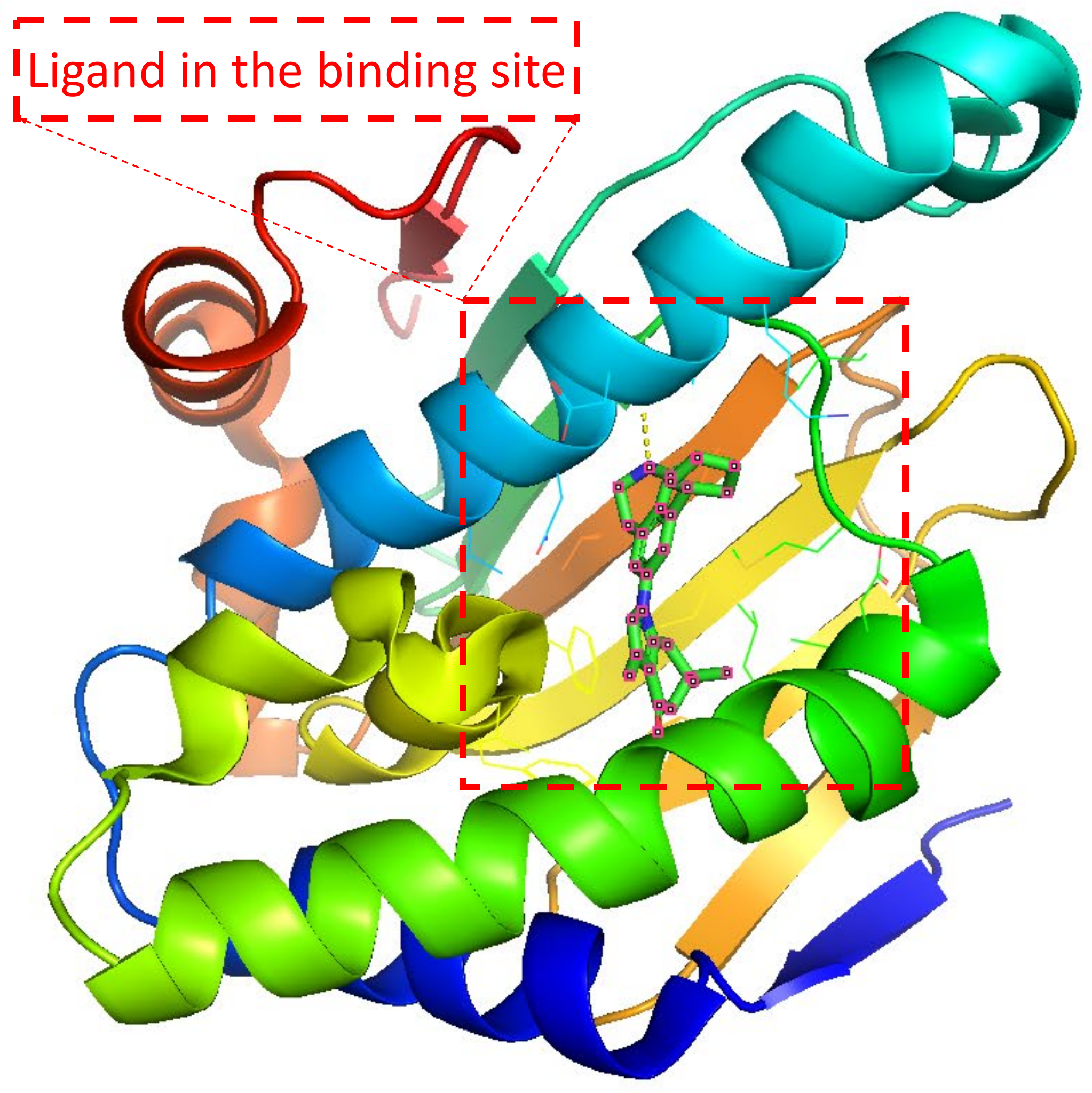}
\caption{Illustration of the protein-ligand pair with PDB ID 4O0B from PDBbind Database. The red dashed square indicates the cartoned binding site and the docked ligand.
}\label{pdb}
\end{figure}

Since the dawn of the genomics era in the 1990s, drug discovery has gone through a transition from a phenotypic approach to a target-based approach \cite{swinney2011were}. Most drug targets encoded by human genomes are complex multimeric proteins whose activities could be modified by binding with drug molecules \cite{overington2006many}. A ligand compound is a substance that forms a complex with the binding site of a protein target, if they are structurally complementary, for therapeutic effects (see Fig. \ref{pdb}). The navigation in the molecule space to find molecular compounds with high binding affinity is called target-specific \emph{de novo} drug discovery.

Traditionally, the ligand was initially identified by screening libraries of commercially available compounds, which are sequentially docked against the protein target. This ligand discovery and optimization process could be time-consuming and resource-consuming with lower probabilities of success \cite{keseru2009influence}. Computational approaches effectively accelerate nearly every stage of drug development.
Most computational approaches are based on generative machine learning models
\cite{molgan, li2022scalable}. However, these generative models hardly work for target-specific drug discovery since they merely learn the molecular distribution.

\begin{figure*}
\centering
\includegraphics[width=16.5cm]{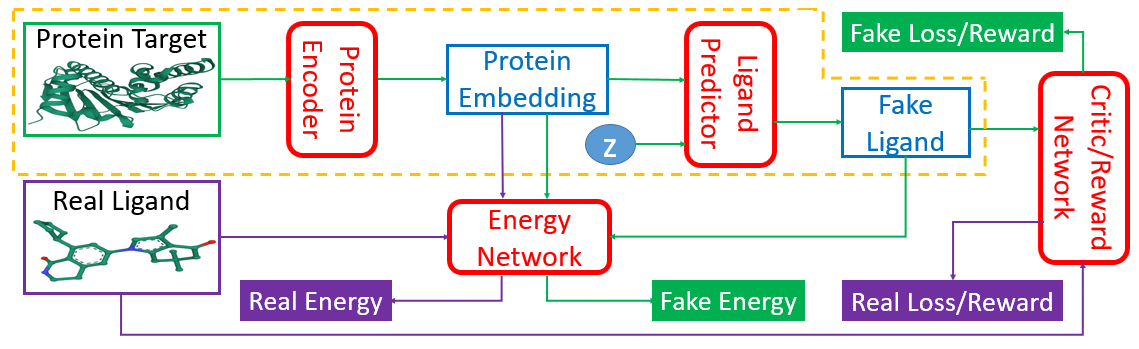}

\caption{TagMol network architecture composed of protein encoder, ligand predictor and two guiding networks. Ligand predictor, a latent-variable predictive model, contains an extra latent variable $z$ sampled from a multivariate Gaussian distribution. Energy network learns using energy differences between \emph{real} and \emph{fake} ligands, and reward network is taught using target reward values, evaluated with an external package RDKit. Blocks with green arrows indicate the generation flow of fake ligands; while blocks with purple arrows indicate real ligand workflow. After training, the network portion within the yellow dashed line can generate ligand candidates for a given protein target. The protein target and real ligand are from the PDB 4O0B pair.}
\label{tagmol}
\end{figure*}

A few computational target-specific approaches also exist, Gupta et al \cite{gupta2018generative} developed a generative RNN-LSTM model to produce valid SMILES strings and fine-tuned the model with drugs with known activities against particular protein targets. Unfortunately, such prior knowledge of protein binders is sometimes unavailable especially for newly identified targets. A recent work in \cite{grechishnikova2021transformer} released this constraint by framing target-specific drug design as a machine translation problem. However, this non-generative model design only provides a probabilistic mapping from targets to ligands, thereby failing to sample ligand candidates for drug targets. CogMol \cite{chenthamarakshan2020cogmol} combined a Variational Autoencoder network and a protein-ligand binding affinity regressor for generating molecules with desired properties. However, the loosely coupled components in CogMol make the sampling less efficient and isn't target-specific. We developed a novel algorithm, \textbf{Ta}rget-specific \textbf{G}eneration of \textbf{Mol}ecules (TagMol), to efficiently sample ligand candidates for drug targets in an end-to-end fashion.

TagMol adopts a protein-ligand binding affinity regressor,
which assigns high energies for ligands incompatible with targets and low energies for those compatible. Thus, our approach falls within the theoretical framework of energy-based models \cite{lecun2006tutorial}. Fig. \ref{tagmol} illustrates the energy-based latent-variable predictive TagMol model which consists of a protein encoder, a ligand predictor (or generator) and two guiding networks. As the latent variable $z$ varies in the multivariate Gaussian distribution, the fake ligand prediction varies over the ligand set compatible with the protein target. The TagMol learning is supervised using discriminator losses
and reward values evaluated from the external cheminformatics package of RDKit. The energy network ensures that generated (or \emph{fake}) ligands are compatible with protein targets, and the reward network guarantees they have desired drug properties.

The contributions of this paper are three-fold: 1) We proposed a novel end-to-end energy-based generative model, TagMol, for target-specific drug discovery; 2) the ligand predictor architecture incorporates an extra latent variable $z$ which entails the generation of ligands with high binding affinity to the input protein target; 3) we implemented graph neural networks with attention mechanism and multiple relations that result in faster and better learning.


\section{Background}
\label{sec:backgroud}

The matching between protein targets and ligands are not unique and are not one-to-one. As reported in \cite{chen2009molecular}, ten drug fragments screened from the ZINC small-molecule database \cite{sterling2015zinc} well inhibited the CTX-M structure, which is a new enzyme family for extended spectrum beta-lactamases.
To exploit the deterministic and probabilistic model design benefits, we devised a latent variable energy-based model for drug discovery.

\subsection{GAN-based Models}

Generative Adversarial Networks (GANs) \cite{gan} are implicitly generative models since they are evaluated using fake sample validity, predicted from a discriminator network. The generator of a GAN is a latent variable model with $z$ being latent variables and $x$ being observed variables. Conditional GAN \cite{mirza2014conditional} is an extended version of GAN which takes any auxiliary information, such as labels, into both the generator and discriminator. 
Based on conditional GAN, Barsoum et al. \cite{barsoum2018hp} developed HP-GAN for probabilistic prediction of 3D human motions based on previous motions.
Latent variables are necessary in modeling biomolecular PDBbind \cite{pdbbind} refined 2017 dataset because the hidden target features, such as protein conformation and cellular localization, explicitly affect the formulation of small-molecule ligands. Based on the conditional GAN, TagMol takes as input the latent variables and protein targets for generating probabilistic ligand candidates for further screening. All possible atoms and bonds in the defined ligand space are assigned with certain probabilities in the generator accordingly. The latent variables would lead the predictions to different sets of plausible ligands conditioned on multiple protein families and conformations.

\begin{figure*}
\centering
\includegraphics[width=16.5cm]{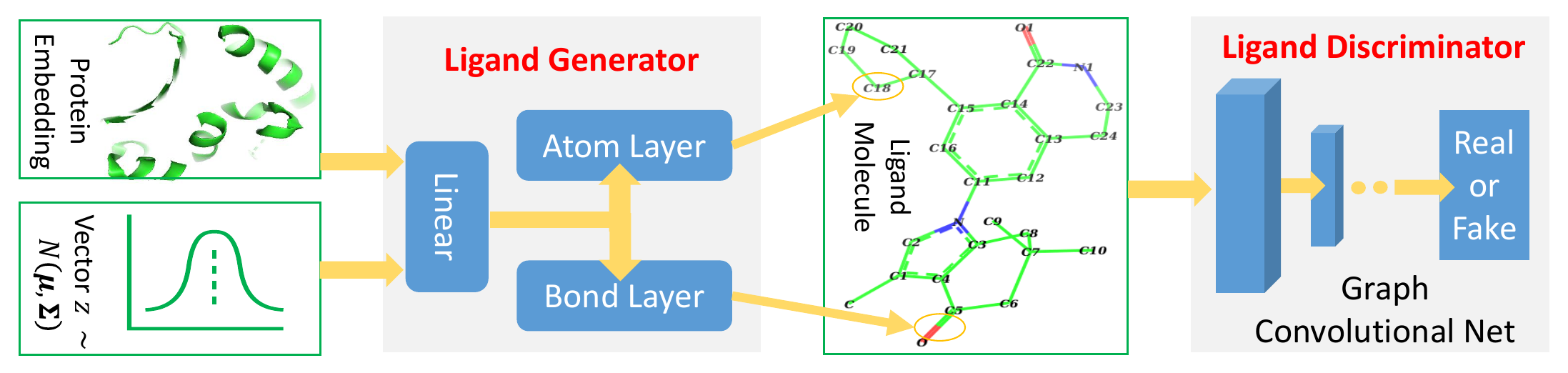}

\caption{TagMol generator and discriminator components for ligand prediction. Protein embedding represents the extracted features from the input protein. A series of linear layers, atom layer and bond layer form the ligand generator. A graph convolutional network forms the ligand discriminator, which assesses the prediction quality with the probability of generated ligand molecules being real.}

\label{cgan}
\end{figure*}

\subsection{Energy-based Models}

Energy-based models (EBMs) \cite{lecun2006tutorial} capture dependencies between variables and evaluate their compatibility by associating a scalar \emph{energy} value. The models are taught by designing an energy function which assigns low energies to correct pairs, and high energies to incorrect pairs. The loss function is designed to measure the quality of the energy function for assigning energy values to different variable pairs during learning and inference. The EBM framework covers a wide range of learning approaches, including probabilistic and deterministic, with respective loss functions. The discriminator in GAN is also an energy-based network which predicts the probability differences (energies) with zeros and ones for fake and real samples, respectively. The energy-based model for probabilistic prediction serves as the proxy for evaluating the binding energy between pairs of protein target and ligand.
While the energy network is probabilistic, protein encoder and reward network parts are deterministic. As for the discriminator in GANs, the critic network in Fig. \ref{tagmol} can also be considered as an energy-based network.

\section{Approach}
\label{sec:approach}

We explain in detail our probabilistic approach for target-specific drug discovery, conditioned on the given protein receptor in this section. The problem is defined as learning the conditional probability of plausible ligands $P(y|x)$, where $y \in \mathcal{Y}$, from a corresponding protein receptor $x \in \mathcal{X}$, given a training set of i.i.d protein-ligand pairs $\mathcal{D}$ $=\{(x_i, y_i)\}_{i=1}^{N_s}$. The ligand space $\mathcal{Y}$ is composed of a bond adjacency matrix space $\mathcal{B}$ $= \{0, 1\}^{N \times N \times B}$ and an atom matrix space $\mathcal{A}$ $= \{0, 1\}^{N \times A}$, where $N$ denotes the maximum number of heavy atoms (excluding Hydrogen) in ligand molecules; $A$ and $B$ represent the numbers of atom types and bond types, respectively.

\subsection{TagMol Algorithm}

TagMol architecture is developed partially based on cGAN (see Algorithm \ref{alg:tagmol}). The generator creates synthetic (or \emph{fake}) data samples from random noises, whereas the discriminator learns to distinguish between the \emph{real} and \emph{fake} samples. The adversarial minimax learning of cGAN is conditioned on extra information, such as class labels. Protein embedding serves as the conditional information in the present study.
As depicted in Fig. \ref{tagmol}, the ligand prediction model takes as input a protein embedding $x$ produced from the protein encoder, plus a latent vector $z$ drawn from a Gaussian distribution. The protein embedding $x$ and vector $z$ are concatenated using early fusion and fed into a series of linear layers, as shown in Fig. \ref{cgan}. The final atom layer and bond layer take the same fused features to generate probable atoms and bonds to form a possible ligand molecule. In our study, molecules are represented using graphs where each node denotes an atom and each edge denotes a bond. The following ligand discriminator (also called critic network since not trained to classify), represented with a Graph Convolutional Network (GCN) (see Fig. \ref{cgan}), evaluates the generation quality. Generator and critic networks are the two major components in TagMol inherited from the GAN architecture.
Apart from the evaluation from the critic network, predicted ligand molecules should be plausible by adding a specific reward network (see Fig. \ref{tagmol}).
Besides, the generated ligands should also exhibit high hit rates when being docked with the provided protein target. To that end, a binding energy network is adopted to enforce target-specific generation. The reward network and energy network are two important guiding networks for predicting plausible and target-specific ligand molecules.

\begin{algorithm}[tb]
\caption{Target-specific Generation of Molecules}
\label{alg:tagmol}
\textbf{Input}: protein-ligand pairs $p_{data}$, iterations $k$, steps $m$\\
\textbf{Parameter}: network parameters $\tau_{Enc}, \phi_G, \psi_D, \theta_E, \omega_R$, and hyper-parameters $\lambda$, $\alpha$, $\beta$, $\gamma$ for loss terms\\
\textbf{Output}: predicted ligands $\hat{y}$
\begin{algorithmic}[1] 
\FOR{$k$ iterations}
\FOR{$m$ steps}
\STATE Sample minibatch of protein-ligand pairs $(x_p, y)$.
\STATE Get embedding from encoder $x \leftarrow Enc_{\tau}(x_p)$.
\STATE Sample minibatch of noise samples $z\sim p(z)$.
\STATE Generate fake ligands $\hat{y}\leftarrow G_\phi(x, z)$.
\STATE $\rhd$ \emph{Update D network parameters.}
\STATE $\psi_D:=\arg\min\limits_\psi \mathcal{L}_D$ $(y, \hat{y}; \psi, \lambda)$ \hfill $\rhd$ \emph{see Eq. \ref{eq:disc}}
\ENDFOR
\STATE Repeat steps 3 to 6.
\STATE $\mathcal{L}_E$ $\leftarrow E_{\theta}(x, y) - E_{\theta}(x, \hat{y}) + \alpha L_2$ \hfill $\rhd$ \emph{see Eq. \ref{eq:le}}
\STATE $\rhd$ \emph{Update E network parameters.}
\STATE $\theta_E:=\arg\min\limits_\theta \mathcal{L}_E$ $(x, y, \hat{y}; \theta)$
\STATE $\mathcal{L}_R$ $\leftarrow \frac{1}{3}[(R_\omega(y)-\text{rdk}(y))^2+(R_\omega(\hat{y})- \text{rdk}(\hat{y}))^2]$
\STATE $\rhd$ \emph{Update R network parameters.}
\STATE $\omega_R:= \arg\min\limits_\omega \mathcal{L}_R$ $(y, \hat{y}; \omega)$
\STATE $\rhd$ \emph{Update G and Enc network parameters.}
\STATE $\phi_G:=\arg\min\limits_\phi(-D(y)+\beta\mathcal{L}_E+\gamma\mathcal{L}_R)$ $\rhd$ \emph{see Eq. \ref{eq:lg}}
\STATE $\tau_{Enc}:=\arg\min\limits_\tau(-D(y)+\beta\mathcal{L}_E+\gamma\mathcal{L}_R)$
\ENDFOR
\STATE \textbf{return} $G_\phi(Enc_\tau(x_p), z)$, \text{for multiple} $z\sim \mathcal{N}$ $(0, 1)$
\end{algorithmic}
\end{algorithm}

\subsection{Ligand Generator and Discriminator}
\label{sec:cgan}

TagMol is taught by first extracting a low-dimensional protein embedding space $\vx$, as shown in Fig. \ref{tagmol}. The objective of the protein encoder $x = Enc_{\tau}(x_p)$ is to extract features associated with the protein binding pocket. A high-dimensional condition makes it hard for the model to build connections between generated ligands and complex proteins. An autoencoder-like unsupervised model learns the latent space representation for all protein targets, rather than the specific binding pockets of interest. Without adopting an autoencoder, the embedding network learns along side all other components in an end-to-end fashion.

As indicated in Algorithm \ref{alg:tagmol}, $G$ denote the ligand generator and $D$ the discriminator. Then the generated (or \emph{fake}) ligand is represented as $\hat{y}=G_\phi(x, z)$ and discriminated with $D_\psi(y)$ where $\phi$ and $\psi$ are learnable parameters in the generator and discriminator networks, respectively. The generator is a feed-forward neural network, after the fusing protein embedding and noise vector. While the discriminator is realized with either a GCN or Graph Attention Network (GAT) \cite{gat} for effectively learning graph representations. The baseline GCN is not specifically described since it is partially explained in GAT whose details are deferred to the following subsection. Each drawn latent vector $z$ creates a plausible ligand molecule with different binding features for a protein target. To prevent GAN training instability, we replace the GAN with WGAN \cite{wgan} for measuring the approximation of generator distribution $q$ to empirical distribution $p$ with Earth Mover (EM) distance. Furthermore, a gradient penalty loss from WGAN-GP \cite{gulrajani2017improved} is adopted to enforce the WGAN Lipschitz constraint.
The ligand generator is trained using WGAN-GP adversarial loss, energy loss and reward loss. 
\begin{align}\label{eq:lg}
	{L}_G = -D(G({x, z}))+\beta \mathcal{L}_{E}+\gamma \mathcal{L}_R
\end{align}
The energy loss measures the docking energy difference between \emph{real} and \emph{fake} ligands. We remark that docking energy here is not computed based on atom interactions in terms of physical force fields, but on an energy function defined in Section \ref{sec:ebm}. The reward loss is calculated using drug properties evaluated from RDKit.

Different from cGAN, the discriminator only takes as input the \emph{real} or \emph{fake} ligands without concatenating the protein target condition. The relation between protein and ligand is guided by the energy network which could be considered as another flexible discriminator. The discriminator loss consists of only the WGAN-GP critic loss.
\begin{align}\label{eq:disc}
	\mathcal{L}_D &= D(G(x,z))-D(y)+\lambda\left(\|\nabla_{\hat{x}}D(\hat{x})\|_2-1\right)^2
\end{align}
where the interpolation $\hat{x} = \epsilon y+(1-\epsilon)G(x, z)$ depends on a uniformly sampled weight $\epsilon \sim U[0, 1]$, and $\lambda$ is a hyperparameter ($\lambda=10$ is used in this study).
The two terms to the left denote the WGAN loss and the right most term denotes the gradient penalty.

\subsection{Energy-based Network}
\label{sec:ebm}

The probabilistic energy-based generative TagMol is developed by estimating the probability distribution $p(y|x)$ over the whole ligand space $\mathcal{Y}$ for a certain protein $x$. Energy network aims to learn an energy function $E_{\theta}(x, y) \in \mathbb{R}$ that attributes low energies to regions near the data manifold $(x, y) \in \mathcal{X}\times \mathcal{Y}$ and high energies to other ligand regions. The energy function defines a probability distribution usually via a Gibbs-Boltzmann density.
\begin{align}\label{eq:gibbs}
	q_\theta(y|x) &= \frac{\text{exp}(-E_{\theta}(x, y))}{Z_\theta(x)}\text{,} \\ Z_\theta(x) &= \int \text{exp}(-E_{\theta}(x, \tilde{y})) \,d\tilde{y}
\end{align}
Where $Z_\theta(x)$ denotes the normalizing partition function. However, $Z_\theta(x)$ is generally intractable due to high dimensionality of target space $\mathcal{Y}$. Unlike Markov Chain Monte Carlo \cite{nijkamp2020anatomy, du2019implicit}, an applicable but inefficient technique to approximate $Z_\theta(x)$, contrastive samples in this study are directly produced from the generator by referring to EBGAN \cite{zhao2017energy}. The gradient of negative log-likelihood loss $\mathcal{L}$$_{energy}$ using contrastive samples is presented below:
\begin{align}
	&\nabla_\theta  {L}_\text{energy}(\theta; p(x, y)) \\
	&= -{E}_{p(x, y)}\left[\nabla_\theta \log q_\theta(y|x) \right] \\
	&= -{E}_{p(x, y)}\left[\nabla_\theta (-E_{\theta}(x, y)-\log Z_\theta(x))\right] \\
	&= {E}_{p(x, y)}\left[\nabla_\theta E_{\theta}(x, y) + \frac{\int \nabla_\theta e^{-E_{\theta}(x, \tilde{y})} \,d\tilde{y}}{Z_\theta(x)} \right] \\
	&= {E}_{p(x, y)}\left[\nabla_\theta E_{\theta}(x, y) - \int \frac{e^{-E_{\theta}(x, \tilde{y})}}{Z_\theta(x)} \nabla_\theta {E_{\theta}(x, \tilde{y})} \,d\tilde{y} \right] \\
	&= {E}_{p(x, y)}\left[\nabla_\theta E_{\theta}(x, y)\right] - {E}_{p(x, y), \tilde{y}\sim q_\theta(\tilde{y}|x)}\left[ \nabla_\theta {E_{\theta}(x, \tilde{y})}\right] \\
	&\approx {E}_{p(x, y), z \sim p(z)}\left[\nabla_\theta E_{\theta}(x, y) - \nabla_\theta E_{\theta}(x, G_\phi(x_e, z))\right] \label{eq:ratio}
\end{align}
where $G_\phi(x_e, z)$ denotes the generated example from noise $z \sim p(z)$ with conditioning on protein embedding $x$. The trick from the last step is that the expectation w.r.t. $\tilde{y}$ is approximated using a single contrastive example $\hat{y}=G_\phi(x_e, z)$ produced from the generator. The loss $\mathcal{L}_\text{energy}$=$E_{\theta}(x, y) - E_{\theta}(x, \hat{y})$ is still an object we want to minimize by pushing down the energies for \emph{real} samples from the dataset and pulling up the energies for \emph{fake} samples. Equation \eqref{eq:ratio} could also be interpreted as minimizing the density ratio between a pair of \emph{fake} and \emph{real} samples such that $Z_\theta(x)$ is bypassed. The final loss function for energy network is defined with an L2 regularization
\begin{align}\label{eq:le}
	{L}_E = {L}_{energy} + \alpha \left(E_{\theta}(x, y)^2+E_{\theta}(x, \hat{y})^2\right).
\end{align}


\subsection{Ligand Reward Network}

Let $R$ denote the reward network for guiding the learning of ligand molecules with desired properties.
The property vector output is then represented as $\vo=R_\omega(y)$ where $\omega$ is the network parameter vector and $y$ is the ligand input. The reward network is the same architecture as the discriminator, except that a property output layer is finally appended rather than an EM distance output layer from WGAN. After a probabilistic distribution of molecules is produced from the generator, a hard categorical sampling step is realized using a straight-through trick for drawing a discretized one-hot ligand molecule represented by a bond matrix $\mathcal{B}$ and an atom matrix $\mathcal{A}$. 

As mentioned in Section \ref{sec:cgan}, a set of GCN or GAT layers are adopted to learn graph-represented molecules by passing node messages iteratively. Bond types convey crucial information in formulating molecules and determine molecule valency validities. Therefore, the relational graph attention network (RGAT) \cite{qin2021relation} is specifically implemented for dynamically learning the importance of edge-specific attribute features. The input to a GAT layer is a molecule graph with $B=|\mathcal{R}|$ relation types and $N$ nodes. The overall input node features are represented with a feature matrix $\vH=[\vh_1, \vh_2, \ldots, \vh_N]^T \in \mathbb{R}$$^{N\times F^\prime}$. The single-head attention coefficient for message passing is defined by incorporating multiple edge relations between $i^{th}$ and $j^{th}$ nodes
\begin{align}\label{eq:attention}
	\alpha_{i, j}^{(r)} &= \frac{\exp{(\sigma(\va_r[\vW_r\vh_i\bigm\Vert\vW_r\vh_j]))}}{\sum_{r^\prime\in\mathcal{R}}\sum_{k\in n_i^{(r^\prime)}} \exp{(\sigma(\va_{r^\prime}[\vW_{r^\prime}\vh_i\bigm\Vert\vW_{r^\prime}\vh_k]))}}
\end{align}
$\forall i: \sum_{r\in\mathcal{R}}\sum_{j\in n_i^{(r)}} \alpha_{i, j}^{(r)} = 1$, where $||$ represents the concatenation operation, $\va_r$ is the $r$-relation weight vector for the attention mechanism, $n_i^{(r)}$ are all first-order neighbors of entity $i$ with relation $r$, and $\sigma$ is the LeakyReLu activation function used throughout our work. The attention weight $\alpha_{i, j}^{(r)}$ can be seen as the contribution from neighbor $j$ to construct output node features $\vh_i^\prime$, after one GAT layer, represented as 
\begin{align}\label{eq:propagation}
	\vh_i^\prime = \sigma\left(\sum_{r\in\mathcal{R}}\sum_{j\in n_i^{(r)}}\alpha_{i, j}^{(r)}\vW_r\vh_j\right)
\end{align}
where $\sigma$ denotes the same LeakyReLu nonlinearity. The attention mechanism produces a single probability distribution over all neighbours of entity $i$ irrespective of relation types. An output feature matrix $\vH^\prime=[\vh_1^\prime, \vh_2^\prime, \ldots, \vh_N^\prime]^T \in {R}^{N\times F^\prime}$ is obtained with higher-order neighbor information. Multiple relational graph attention layers could be applied for learning better graph representation. 

For ligand property prediction, graph-level features are retrieved by referring to the graph aggregation method \cite{rgat} which concatenates the mean of node representations with the feature-wise maximum across all nodes.
\begin{align}\label{eq:aggregation}
	\vg(\vH^\prime) = \left(\frac{1}{N}\sum_{i=1}^{N}\vh_i^\prime\right)\bigg\Vert\left[\bigoplus_{f=1}^{F^\prime}\max_{i}\vh_{i, f}^\prime\right]
\end{align}
Where $\bigoplus$ denotes the element-level concatenation of feature maxima across nodes. A final fully connected layer is added for producing predicted drug property vector rew$(y)$. Properties evaluated from RDKit are taken as target values rdk$(y)$ for \emph{real} and \emph{fake} ligands. The reward loss function is defined with MSE loss between predicted and target properties as shown in Algorithm \ref{alg:tagmol}.

\section{Experiments and Results}


\subsection{Dataset and Metrics}

All the experiments are conducted with the biomolecular PDBbind \cite{pdbbind} refined 2017 dataset which contains 3843 and 663 molecules in training and testing sets, respectively. Ligands with more than 32 heavy atoms are trimmed by removing the atoms with a small number of bonds with the neighboring atoms. Heavy atom types include carbon, nitrogen, oxygen, fluorine, sulfur, and chlorine.

Learning results of the proposed CGAN-based models are evaluated using Fréchet distance (FD) which estimates the similarities between generated ligands and \emph{real} ones. To evaluate the effectiveness of a protein target, variants without protein embedding ($x\_dim=0$) are trained as well for comparison. Each sample batch of real or fake molecules is concatenated to a multidimensional point in the sampling distribution. Both atom and bond features from sampled molecules are considered for FD score calculation by referring to \cite{li2021quantum}. The performance from a non-conditional model without protein embedding serves as the FD baseline to evaluate the energy-based generative models.

\subsection{Implementation Details}

Initially the reward network and energy network are dropped for conducting the ablation study on protein embedding dimension. All GAN variants are trained with a minibatch of 64 molecules with the Adam optimizer on a single RTX 2080Ti GPU. The learning rate is initially set to 1e-4 and updated to 1e-5 after 200 training epochs. All models are trained with 1000 epochs, and early stopping is applied if the learning diverges.

\begin{figure*}
\centering
\includegraphics[width=\linewidth]{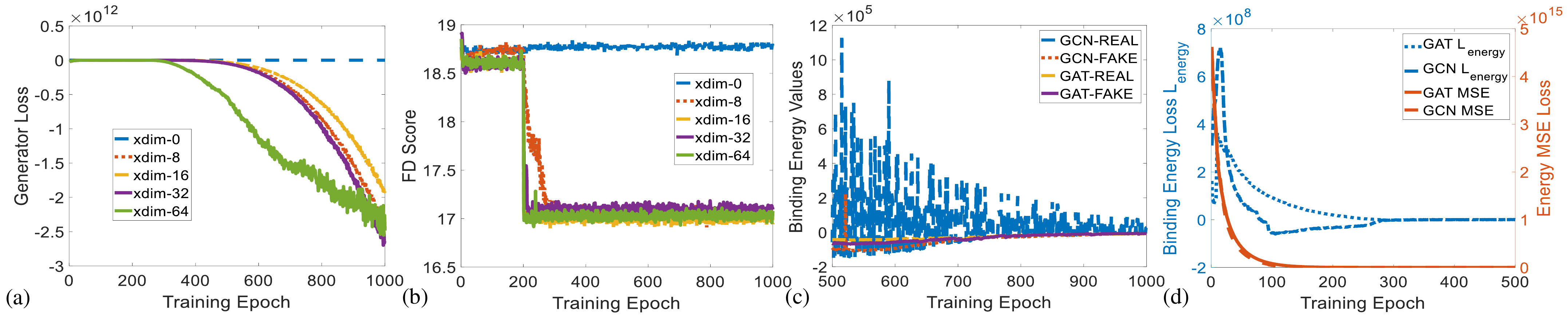}
\caption{Experiment setup on protein embedding dimension and binding affinity scores for GCN- and GAT-based TagMol models. (a) Training generator loss considerably goes down at epoch 200 where learning rate was updated; (b) FD score for xdim 16 shows slightly better than other non-zero dimensions; (c) faster and stable learning is observed for TagMol with GAT layes; (d) Similar binding affinities are achieved between the target pairs with \emph{real} and \emph{fake} ligands. Learning rates were set to 1e-5 for evaluating the overall TagMol model in (c-d).}
\label{results}
\end{figure*}

\subsection{Ablation Study}

The protein embedding dimension (xdim) affects the cGAN performance in terms of generator loss. When xdim is set to zero, ligand generation is independent of the given protein, indicating a non-conditional model. Therefore, the binding pocket in the target cannot guide the target-specific drug discovery, as shown in Fig. \ref{results}(b) where the FD score hardly decreases. When xdim is large, the model is more complex which corresponds to a steeper generator loss curve (see Fig. \ref{results}(a)). However, the variance effect caused by Gaussian noise $z$ is mitigated. The variance is beneficial since various ligands could be created for a certain target. Embedding dimensions ranging from 0 to 64 were tested for finding a suitable dimension that achieves a better FD score. A protein embedding dimension of 16 was selected through the ablation study on it. It is worth noting that the sudden changes in generator loss and FD score were caused by the learning rate decay at milestone epoch 200. All following experiments are conducted with 16 dimensional protein embedding.

\subsection{Results}

The energy-based network $E_{\theta}(x, y)$ reflects the final binding affinity between protein target and ligand candidates. The learning quality of generated ligands are thus evaluated using binding energy loss $\mathcal{L}_\text{energy}$ and scaled MSE loss $\alpha \left(E_{\theta}(x, y)^2+E_{\theta}(x, \hat{y})^2\right)$. Hyperparameter $\alpha$ was set to $1e-3$ after several rounds of warm-up learning. As mentioned earlier, two types of GNN layers, \emph{i.e.} GCN and GAT, were tested for comparing the binding energy values. TagMol results with these two settings were plotted in Fig. \ref{results}. We remark that, in panel (d), the final negative $\mathcal{L}_\text{energy}$ value reveals a better affinity for \emph{real} ligands. The value eventually comes close to zero, which indicates the \emph{fake} ligands have a similar binding affinity relative to \emph{real} ones. The right y axis displays the scaled MSE losses for GCN- and GAT-based energy models. A slightly lower MSE loss was observed for GCN models for the first few dozen epochs. $\mathcal{L}_\text{energy}$ and MSE curves become less distinguishable after learning 500 epochs.

To compare GCN and GAT in detail, we plotted the learning variance (or instability) of GCN-based energy models in Fig. \ref{results}(c). The baseline GCN models showed relatively unstable curves due to the lack of an attention mechanism. The fluctuating loss for \emph{real} ligands is possibly attributed to the bad GCN early-stage learning quality such that each weight update causes large binding energy changes. Binding energy values of GAT for \emph{real} molecules turned out to be smaller than \emph{fake} ones after 875 epochs. However, energy scores corresponding to GCN-REAL are still slightly higher than \emph{fake} counterparts after 1000 training epochs. This is another indicator of the advantage GAT layers have over GCN layers. We remark that the predicted energies provide the proxy for indirectly evaluating binding affinities, rather than physically evaluate the compatibility between protein-ligand pairs in terms of physical force fields, because of the log-likelihood loss function in EBMs.

\section{Conclusion}

We proposed a probabilistic energy-based model called TagMol for target-specific drug discovery. The model specifically evaluates the binding affinity scores between protein-ligand pairs with an EBM.
The protein embedding dimensions were tuned within the cGAN components first. Generated ligands achieved comparable binding energy scores for TagMol models with GAN and GAT layers. However, a faster and more stable learning is observed for GAT layers with attention over all atoms in drug molecules.

\vspace{12pt}
\bibliographystyle{IEEEtran}
\bibliography{IEEEabrv,ref}

\end{document}